\documentclass[runningheads]{llncs}
\usepackage[T1]{fontenc}
\usepackage{graphicx}
\usepackage{graphicx}
\usepackage{amsmath}

\usepackage{amsthm}
\usepackage{booktabs}
\usepackage{algorithm}
\usepackage{algorithmic}
\usepackage{caption,tabularx,booktabs}
\usepackage{dblfloatfix}
\usepackage{makecell}
\usepackage{multirow}
\newcolumntype{Y}{>{\raggedright\arraybackslash}X} 
\usepackage{arydshln}

\begin{document}
\title{Detecting AI Assistance in Abstract Complex Tasks}
%
%
\author{Tyler King\inst{1}\orcidID{0000-0003-4394-950X}, Nikolos Gurney\inst{2}\orcidID{0000-0003-3479-2037}, John H. Miller\inst{3,4}, \and Volkan Ustun\inst{2}} 
\authorrunning{T. King, N. Gurney, J. Miller, and V. Ustun}

\institute{Cornell University, Ithaca NY 14850, USA \\
\email{ttk22@cornell.edu} \and
University of Southern California, Los Angeles CA 90094, USA\\
\email{gurney@ict.usc.edu, ustun@ict.usc.edu} \and
Carnegie Mellon University, Pittsburgh PA 15213, USA \\
\email{JM7T@andrew.cmu.edu} \and
Santa Fe Institute, Santa Fe NM 87501, USA}
\maketitle              
\begin{abstract}
  Detecting assistance from artificial intelligence is increasingly important as they become ubiquitous across complex tasks such as text generation, medical diagnosis, and autonomous driving. Aid detection is challenging for humans, especially when looking at abstract task data. Artificial neural networks excel at classification thanks to their ability to quickly learn from and process large amounts of data---assuming appropriate preprocessing. We posit detecting help from AI as a classification task for such models. Much of the research in this space examines the classification of complex but concrete data classes, such as images. Many AI assistance detection scenarios, however, result in data that is not machine learning-friendly. We demonstrate that common models can effectively classify such data when it is appropriately preprocessed. To do so, we construct four distinct neural network-friendly image formulations along with an additional time-series formulation that explicitly encodes the exploration/exploitation of users, which allows for generalizability to other abstract tasks. We benchmark the quality of each image formulation across three classical deep learning architectures, along with a parallel CNN-RNN architecture that leverages the additional time series to maximize testing performance, showcasing the importance of encoding temporal and spatial quantities for detecting AI aid in abstract tasks. 
\end{abstract}
\section{Introduction}

The proliferation of highly-capable AI systems that help people complete complex tasks has underscored an anticipated but underappreciated technical gap: the ability to detect when a person worked with or used an AI to complete a complex task (e.g., \cite{cotton2023chatting,floridi2020gpt,mitchell2023detectgpt}). Complex tasks are those characterized by multiple choice variables that share non-linear interactions \cite{gurney2023role}. These include mundane activities, such as grocery shopping, and more notable tasks, such as developing a new software program, routing problems, and protein folding. We document how to use off-the-shelf deep learning models to detect whether a person worked with an AI helper during a complex task. Important features of our approach include the ability to detect AI assistance from data that a person cannot use to accomplish the same, much better-than-chance performance on a relatively small dataset, and the ability to reason about a task that is abstract---which may suggest generalizability. 

Many deep learning use cases, such as large language models for text prediction, are conceptually approachable for humans, resulting in easy misuse. Moreover, people are capable of reading a text and guessing where it falls on the spectrum from completely human-generated to completely AI-generated, although their accuracy in doing so is often poor \cite{uchendu2023understanding}. This shortcoming motivates the need for detection systems. Researchers and institutions are thus devoting considerable resources to automated detection of AI help, such as content generated by large language models \cite{cotton2023chatting,mitchell2023detectgpt,rodriguez2022cross}. However, other deep learning use cases, such as complex routing problems or protein folding, are not conceptually approachable for humans. Detecting input from AI helpers in such use cases is still valuable (arguably more), albeit understudied. Fortunately, as we demonstrate, detecting help from AI in abstract, complex tasks is not meaningfully more difficult. We show that our dataset construction naturally extends to LLMs and multi-agent systems, where exploration/exploitation heuristics are common \cite{murthy2024rex,leonardos2022exploration} and thus can be leveraged as a tool to classify AI aid.  

We rely on data from an experiment that explored human subjects' ability to complete a complex choice task solo and with an AI helper \cite{gurney2022experimental,gurney2023role,gurney2023aiTeammate}. The experimental task asks participants to tune on-screen dials, similar to those of a transistor radio, to discover an optimal setting. Each dial in the task gives participants access to a unique dimension of an n-dimensional problem; in the case of the data we analyze, participants explored 3-dimensional spaces. A 3-dimensional problem requires two dials: if the $X$ and $Y$ dimensions are accessible by the dials, then feedback is given about the $Z$ dimension. The number of dials and the linearity of their relationships determines the task complexity. While perfectly linear systems can be easily optimized by following the gradients along each dimension, finding the optimum in a nonlinear system is much more challenging since optimizing one dial facilitates finding one of many possible optima. 

To illustrate, consider the task of finding the highest peak in the Japanese Yamanashi Prefecture: Mount Fuji. The landscape is such that, even in a dense fog, an individual who simply follows an uphill gradient will eventually summit the peak. Now contrast that with Pennington County, in the U.S. state of South Dakota, i.e., the home of Bad Lands National Park. Even without a dense fog obscuring their vision, an explorer will likely find it difficult, if not impossible, to locate the highest peak in the park by simply following gradients. Both searches can be thought of as 3-dimensional tasks. Cast as a dial tuning task, one dial would allow a person to explore the east-west and the other the north-south dimension of the landscapes. Feedback in the form of elevation for any given east-west and north-south combination would be the third dimension that a person tries to optimize. 

Each participant in the data we studied did four tasks, two alone and two with an AI helper. In both instances, participants searched a ``simple'' (Mount Fuji) and ``complex'' (Badlands) landscape in random order using two on-screen dials (Figure \ref{figure:complete_model}(a)). The landscapes were procedurally drawn and unique, meaning every landscape for every participant was unique. They were not given specifics about the task other than they were to try and optimize the dial settings. Each time they submitted a dial setting, they got feedback revealing the ``elevation'' of the landscape at that particular location. They were allowed to break off search on a landscape at any point. The experimental design crossed gain-loss utility with an anchoring treatment. Participants in the loss (gain) frame were trying to not lose (earn) points. Participants in the anchor condition knew the best possible value for each task they completed. Details about the individual and team effort outcomes are available in \cite{gurney2023role,gurney2023aiTeammate}. 

We project participants' search efforts into images by using the $X$ dimensional dial for image width, the $Y$ dimensional dial for image height, and the $Z$ dimension (the dial combination value) to determine pixel value. The algorithm that generated the landscapes also discretized them such that each dial had 24 unique settings. We use the same discretized values (since it is what participants saw) to generate our images, meaning our dataset consists of 24x24 pixel images. We also generated richer image formulations by encoding metadata such as exploration/exploitation movements and selected locations (as determined by the dials) in additional image channels. We test these images on various deep neural network architectures, including a modified LeNet-5 implementation (to maintain the original parameter count), ResNet-18, and SB-ResNet-18, a modified ResNet-18 with a single residual block. Each model significantly outperforms random chance, indicating the potential of identifying AI aid in abstract, complex tasks. Furthermore, our results indicate that encoding exploration and exploitation states into images improves performance, along with empirically showing smaller model architectures such as SB-ResNet-18 and LeNet-5 outperform larger models such as ResNet-18 across all image formulations due to their better-parameterized regime. 

\section{Related Work}

The need to differentiate between human and AI effort, although long foreshadowed by the likes of the Turing test \cite{turing1950computing}, is admittedly a very contemporary problem \cite{nishihata23hai}. Despite the monumental accomplishments of Deep Blue \cite{campbell2002deep}, AlphaGo \cite{silver2016mastering,silver2017mastering}, and other AI systems that have learned to outperform or mimic humans (e.g., \cite{hannun2019cardiologist,vinyals2019grandmaster,schrittwieser2020mastering}), their behavior has always been decidedly \textit{machine}. Although a lay observer may not have the ability to discern between AlphaGo and a world champion Go player---a thought that conjures \textit{ELIZA} \cite{weizenbaum1966eliza}---an informed one almost certainly can and possibly even learn tricks to defeat these systems \cite{wang2022adversarial}. 

Recent advances in large language models (LLMs) point to the need for a way to differentiate between human-only, AI-only, and AI-aided performance \cite{thorp2023chatgpt,van2023chatgpt}. A number of clever methods are now available for telling whether an AI, such as OpenAI's GPT, or a human-generated a document. DetectGPT, for example, is a zero-shot method that relies on the log probabilities from the model in question and perturbations of a sample text to estimate whether GPT (or a similar language model) generated it \cite{mitchell2023detectgpt}. In another example, researchers used a reasonably small set of labeled data from subject matter experts to train classifier models capable of identifying whether an earlier version of OpenAI's GPT or a human-generated certain texts \cite{rodriguez2022cross}. Importantly, the classifier worked outside of distribution, meaning it could correctly label text samples from biomedical research even though its training dataset only consisted of labeled physics data. Despite this recent focus on classifying AI aid in text generation, there is a body of problems where detecting AI aid could be necessary \cite{Ponti2022aiaid,patil22hai}, particularly those that can be classified as abstract tasks \cite{gurney2023role}.

Our method, rather than relying on features of the model that generated the data (\cite{mitchell2023detectgpt}) or subject matter experts (\cite{rodriguez2022cross}), is to simply convert data into a format that well-known deep learning architectures can readily process. We focus on two popular architectures: LeNet-5, one of the first neural networks to use convolutional operations \cite{lecun1998lenet}, and ResNet-18, one of the first architectures to allow the construction of deep neural networks by preventing vanishing and exploding gradients on deeper layers \cite{he2015resnet}. Critically, we do not need to know any specifics about how the AI assistant functioned to detect that help was given, just basic insights into human behavior (such as how humans perform tradeoffs between exploitation and exploitation) to inform data formulation. 

\section{Dataset Formulation}

\begin{figure}[t]
\begin{center}
\includegraphics[width=0.975\linewidth]{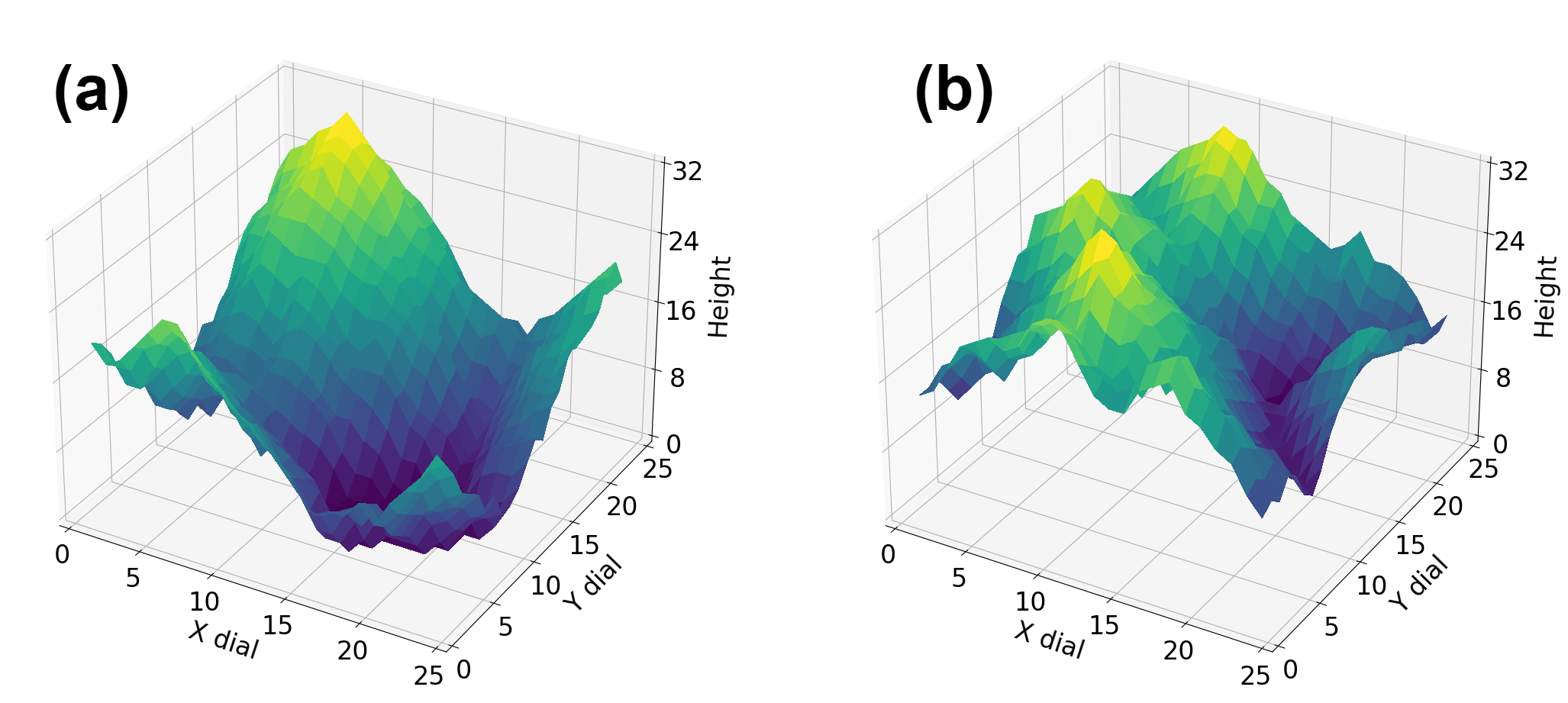}
\caption{Randomly selected example of 1-peak \textbf{(a)} and 4-peak \textbf{(b)} height maps taken from the raw dataset. Note that the single peak environment is easier to optimize for a human agent, ultimately resulting in a task where classifying AI aid is harder. } \label{figure:height_map}

\end{center}
\end{figure}

We lifted the data to test our method from a set of papers exploring how people make complex choices alone and with an AI helper \cite{gurney2022experimental,gurney2023role,gurney2023aiTeammate}. The study participants used two on-screen dials to search procedurally generated landscapes for the highest point in each (one dial for the $X$ and another for the $Y$ dimension). They did not see a visualization 
of the landscapes or receive an explanation of the task grounding it in a landscape search. The experiment relied on a simple incentive structure in which participants' pay was positively correlated with the dial setting value. In other words, they earned the most when they found the global maximum. Our AI assistance detection method requires translating the landscape and behavioral data into image matrices. 

The original authors discretized each landscape into a $24 \times 24$ grid, meaning that a given landscape contains 576 locations with associated elevation values. Landscapes had one or four peaks. Examples of each are in figure \ref{figure:height_map}. The four-peaked landscapes always had a unique global maximum and three unique local maxima. The algorithm that generated the landscapes ensured that each peak had prominence, meaning a valley was between any two peaks. It then smoothed the topology according to a set of adjustable parameters. An important feature of the landscapes is that they are continuous over the edges, meaning that exiting a landscape at the maximum of the $X$ dimension will lead one back to the minimum of the $X$ dimension. Thus, rolling the two ends of the $X$ dimension together will form a tube with consistent topological features. The same is true for the $Y$ dimension; rolling both dimensions results in a torus. Although the algorithm generated each landscape such that elevations were always on the [0,32] interval, the experimental design included a perturbation of the heights to reduce context effects. 
We use the underlying values (those generated by the algorithm before perturbation) in our image matrices.

The image matrices we use are projections of the basic $24\times 24$ landscape matrix. Each location on a landscape---a dial setting combination, i.e., node---is a height value and a matrix of indicators for whether a participant visited a given node. We augmented this basic data with information about experimental conditions and theoretical propositions about how people solve complex choices. 

We created four image matrices to convert raw data into image embeddings:
\begin{enumerate}
    \item Sharp image matrix (sharpIM)
    \item Smooth image matrix (smoothIM)
    \item Basic multi-channel image matrix (bmcIM)
    \item Complex multi-channel image matrix (cmcIM)
\end{enumerate}

\begin{figure*}[!t]
\begin{center}
\includegraphics[width=0.925\linewidth]{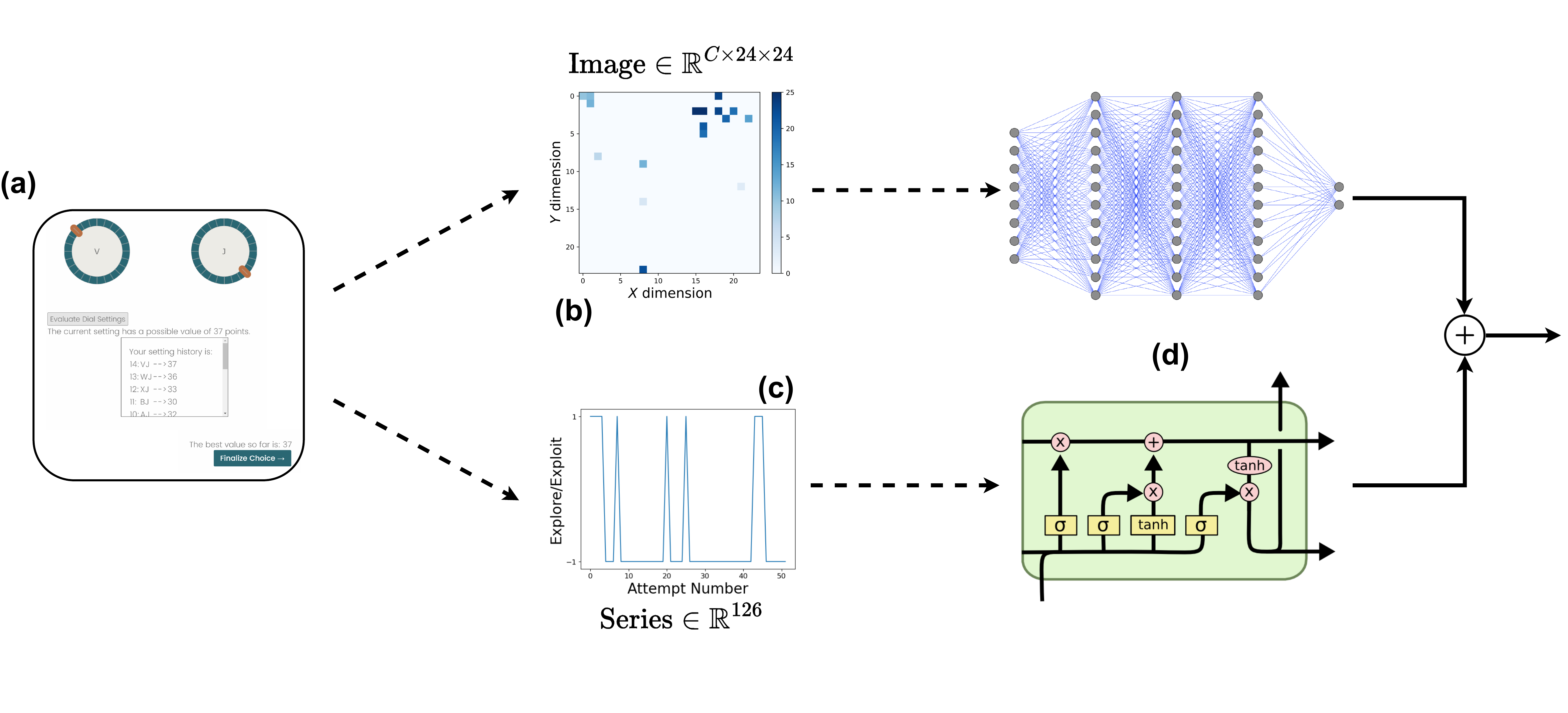}
\caption{Proposed model architecture with supplemental LSTM encoding explore/exploit states. \textbf{(a)} a screenshot from \protect\cite{gurney2023aiTeammate}, showing a solo (i.e. no AI helper) effort where a player adjusts the left dial (the $X$ dimension) and the right dial ($Y$ dimension) to try and find the maximum elevation ($Z$ dimension value). \textbf{(b)} SharpIM image representation extracted from the raw data. 
$C$ is dependent on the image formulation. 
\textbf{(c)} Series representation of exploration and exploitation states. Series lengths less than 126 are padded with 0s to maintain length. \textbf{(d)} Neural network architectures for the image and series data, which are then concatenated before being fed through an MLP. LSTM architecture (bottom) is from \cite{olah2015lstmvis}.    } \label{figure:complete_model}
\end{center}
\end{figure*}

SharpIM (Figure \ref{figure:complete_model}(b)) consists of a single channel where non-selected nodes have no weight and selected nodes have their weight determined by the selected node on the height map. For example, if a given node has a height of 25, then its corresponding weight in sharpIM is 25. This image formulation accounts for a participant's decisions (which dial combinations to check) and the underlying height map, thus serving as a simple baseline to benchmark alternative preprocessing approaches.


Explored heights are sparse in most images, thus we created smoothIM by applying a $3 \times 3$ convolutional filter over sharpIM where convolutions on edge pixels 
connect to pixels on the other side of the 
landscape dimension to soften their effect. This wrapping follows the natural landscape pattern, as the height map is continuous when traversing across opposing edges (north-south and east-west). 
It also, arguably, can represent a participant's intuition for the consistency of location values, provided they conducted sufficient fine-tuning of the dials to learn that the landscapes were locally consistent. 

We hypothesized that humans and computational intelligence differ in how they make explore-exploit decisions, which humans often find challenging (i.e. determining whether an explore or exploit decision is optimal at a particular time \cite{wilson2021balancing}). Exploitation, as defined in the original work, consists of checking locations that are a Manhattan distance of two or less from a previously checked location. Exploration is defined as Manhattan distance movements of three or more from a previously visited location. If human and AI explore-exploit decisions generate different variance structures, then we can use it to look for AI input. This property arises in modern LLM frameworks \cite{murthy2024rex} where exploration-exploitation tradeoffs are integrated into the model's action space, along with multi-agent learning where learned agents may converge optimally with respect to an exploration-exploitation tradeoff \cite{leonardos2022exploration}.

With that knowledge, we formulated bmcIM as a three-channel setup to leverage any such difference. The first channel is the height map as a $24 \times 24$ layer, the second is a binary layer for whether a node was visited (1) or not (0), and the third is a ternary layer coding for exploration (1), exploitation (-1), and unselected (0). These channels may allow the models to learn the same information in sharpIM while detecting whether a movement was an explore or exploit decision. We anticipate that this type of decision-making is one in which humans and computational intelligence differ significantly. As a result, humans and AI will generate detectable different variance structures, which we exploit when looking for AI input. 

Finally, we formulated a five-channel setup, cmcIM, consisting of the bmcIM layers along with two binary layers that track explore states (Manhattan distance $\geq 3$). One layer is 1 if a move is a horizontal exploration (left dial) and 0 otherwise, while the other layer is 1 if a move is a vertical exploration (right dial) and 0 otherwise. These channels encourage the model to focus on explore states, which the data suggest were more erratic with the AI helper.

Additionally, we introduce an auxiliary time-series dataset that explicitly encodes exploration and exploitation states of players as a sequential task, capturing the temporal nature of decision-making as players work through the provided task. Exploration states were encoded as 1s, while exploitation states were encoded as -1 (similar to bmcIM and cmcIM), and were padded with additional 0s to ensure correct shapes. This serves as supplemental data for the core image matrices due to the smaller data size lacking the same expressivity of the various image formulations.

\section{Our Approach}

\subsection{Model Architectures}
By converting raw data into images and time series, we can leverage tools developed for image classification
, such as convolutional neural networks (CNNs), residual neural networks (ResNets), and LSTMs \cite{rawat2017dcnn}. We chose three neural network models for images: LeNet-5 \cite{lecun1998lenet}, ResNet-18 \cite{he2015resnet}, and a modified version of ResNet-18 with a single residual block (SB-ResNet-18).
We used SB-ResNet-18 to take advantage of its slip connections while avoiding overfitting on the smaller dataset \cite{philipp2018gradients,yun2019resnetsbetter}. It also results in a markedly smaller model \ref{table:parameters}.

Residual networks (such as ResNet-18) 
were developed to counter concerns with vanishing and exploding gradients, along with helping train larger neural networks \cite{he2015resnet} via slip connections that smoothen loss landscapes \cite{yun2019resnetsbetter,shamir2018resnets}, helping generalization error and trainability \cite{li2018visualizing}. LeNet-5 is another commonly tested architecture that utilizes convolutional operations to learn local patterns in an image \cite{lecun1998lenet}, which was of interest both for this property and its low parameter count that matched our smaller datasets.

LSTMs \cite{hochreiter1997lstm} have shown promising performance in time-series tasks, particularly with small data and dataset size \cite{ezencan2020comparison}. As a result, we chose to utilize this architecture over more modern approaches such as transformers \cite{vaswani2023attention}. 

Finally, we introduce a parallel CNN-RNN architecture inspired by prior work \cite{Yao2019cnnrnn} that concatenates outputs from an image model and an LSTM. The complete architecture can be seen in Figure 1. While heavy-duty model search methods, such as neural architecture search, exist and could 
optimize model selection, we felt that familiar models best demonstrate our concept. 
These models are not only simple but also sufficient to showcase the identification of AI assistance by comparing various parameterization regimes \cite{kyriakides2020nas}.

Note that we do not show results solely for LSTM due to poor performance. All three image model architectures (LeNet-5, ResNet-18, SB-ResNet-18), along with the supplemental LSTMs were implemented in Pytorch \cite{paszke2019pytorch}. 

\begin{enumerate}
    \item \textbf{LeNet-5}: Our implementation utilizes the same order of layers as outlined in LeCun's seminal paper \cite{lecun1998lenet} with three additional ReLU layers that do not impact the total parameter count. Our network is as follows: 5 by 5 convolution to downsample to a 20 by 20 image with 12 channels, then we apply a ReLU over our weights, and then max pool with a 2 by 2 kernel with a 2 by 2 stride to downsample to a 10 by 10 image. We repeat this process, again convolving with a 5 by 5 filter to obtain a 6 by 6 image with 30 channels, applying a ReLU layer, and then a max pool layer with a 2 by 2 kernel and 2 by 2 stride to downsample to a 2 by 2 image. We then flatten the layer and have a fully connected layer with an output of 180 neurons, which is connected by a second fully connected layer to 2 neurons, and then the softmax is taken. 
    \item \textbf{ResNet-18}: We utilize the same framework of ResNet-18 proposed in He et al. with the exception that our input images have the dimensions of 24 by 24 and the input channel count varies depending on image formulation \cite{he2015resnet}. We also modified output to be a binary layer instead of the standard 1000-way classification.
    \item \textbf{SB-ResNet-18}: To lower the total parameter count of ResNet-18 (which we believe is overparameterized for our datasets), we flatten output after the first residual block and add a fully connected layer to 2 neurons that represent our binary classification task. 
\end{enumerate}

\begin{table}[h]
\centering
\begin{tabular}{cccc}
\toprule
Model & 1-channel & 3-channel & 5-channel \\ \midrule
LeNet-5 & 58,484 & 59,084 & 59,684  \\
ResNet-18 & 11,683,240 & 11,689,512 & 11,695,784 \\
SB-ResNet-18 & 151,362 & 157,634 & 163,906 \\ \bottomrule
\end{tabular}
\caption{Parameter count for all architectures and all image formulations. SharpIM and smoothIM are both one-channel images, while bmcIM has three channels and cmcIM has five channels. Parameter count across image formulations are similar and only vary in the first convolutional layer.}
\label{table:parameters}
\end{table}

\subsection{Data Preparation}

Each of the 398 Amazon Mechanical Turk (AMT) workers (i.e., participants) attempted the 1-peak and 4-peak environment with and without the aid of AI, granting access to $398 \times 4$ distinct data points, along with guaranteeing that the dataset was balanced. To avoid values in the height map from dominating model classification, we applied a normalization scheme across each channel of the image formulations. 

A small number of participants were clear behavioral outliers---one, for example, checked every location in a landscape. To alleviate the impact of such behavior without being biased in our approach, we removed 2.5\% of trials from each tail from the median user interaction (defined as the number of sampled locations), keeping $\approx$ two standard deviations (above and below the median) worth of data. 
The skew is so pronounced that two standard deviations from the mean would remove only trials with high user interaction while effectively disregarding low effort trials. Obviously, both types of outlying behavior will hamper the ability to achieve reasonable model fits.  

\section{Training}

We performed a random 80/20 dataset split, where we used 80\% of the data for training and 20\% for testing. All experiments were run on an Nvidia RTX 8000 GPU. We examined the data holistically and by the peak count. We label the datasets accordingly: x1 (x4) refers to data from participants' solo and AI-assisted search of 1-peak (4-peak) landscapes; \textit{all} refers to the combination of both 1-peak and 4-peak data. We trained multi-channel models past 99\% training accuracy on the complete post-processed dataset, but due to sharpIM and smoothIM being simpler models training was stopped around 97-98\%. To achieve this accuracy, ResNet-18 was trained for 25 epochs, SB-ResNet-18 was trained for 45 epochs, and LeNet-5 was trained for 110 epochs. The best single trial (defined as a fully trained neural network) accuracy achieved and 100 trial average (which was averaged epoch-wise) was taken for all four model formulations on all three models with either just the 1-peak dataset, 4-peak dataset, or the combined dataset.

Due to low-effort participants forcing a noisy, heavy-tailed distribution in our dataset, we decided to use Adam as our optimizer over classical small dataset optimizers such as gradient descent or stochastic gradient descent (SGD) \cite{kingma2014arxiv,zhang2020adamvssgd}. We made this decision because the data features can inhibit methods, such as SGD, by forcing underfitting. Using Adam removed some of the focus on hyperparameter tuning since it is fairly robust to poor hyperparameter selection \cite{Goodfellow-et-al-2016}. Batch normalization was utilized in lieu of dropout (for ResNets) due to concerns with reducing accuracy stemming from inaccurate dropout neuron selection \cite{garbin2020bnvsdrop}. 

The noisy nature of our dataset required us to use an averaging technique. Without averaging, we would benchmark stationary 80/20 splits that could unequally benefit certain image formulations and model architectures. To counter this, we averaged testing results and loss over 100 trials with randomized data splits, leveraging the law of large numbers to obtain the expected performance of each model architecture and image formulation combination. This was also why we did not use ensemble learning methods, as trying to optimize testing metrics may not transfer to a hidden testing set. 

Furthermore, we averaged performance epoch-wise (for every epoch, we average the classification performance of all testing results) as opposed to averaging trial-wise best performance due to high variance from epoch to epoch during testing. This means that reporting best performance averaged trial-wise, as is common, may result in models with high variance. By looking at epoch-wise averaged results, however, we can assess a reasonable early stopping condition. Note that reporting average and standard deviation of validation accuracies epoch-wise is a standard approach for small datasets \cite{xu2018how,cai21graphnorm}.

\begin{figure*}[!b]
\begin{center}
\includegraphics[width=\textwidth]{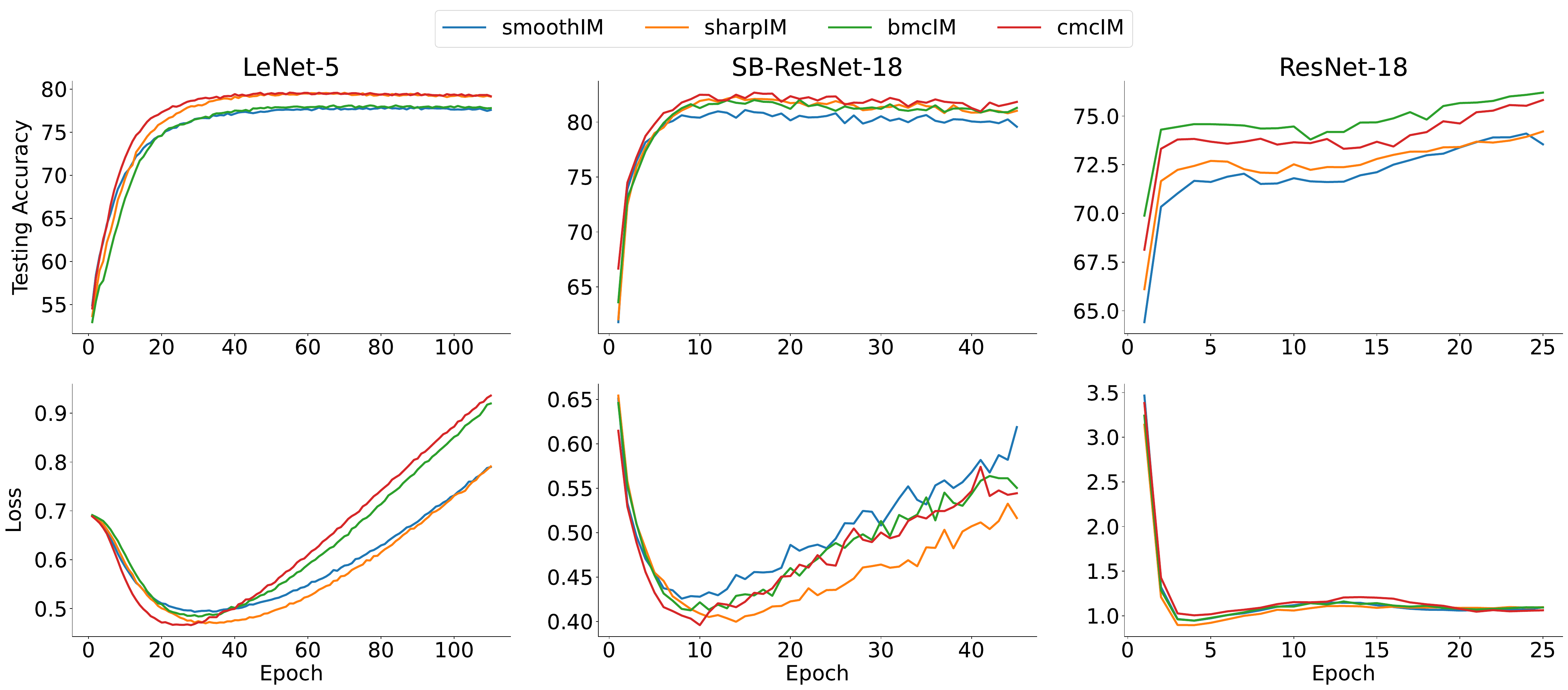}
\caption{Testing accuracy and loss of each model architecture on all image formulations. ResNet-18 was indicative of double descent, indicating that loss bottoms out initially and then increases before decreasing and plateauing. The other two architectures achieved typical loss curves that are accounted for by bias-variance trade-offs, where loss is minimized when bias and variance error are both low. SB-ResNet-18 achieved the best testing accuracy on most image formulations, followed by LeNet-5 and then ResNet-18. } \label{figure:results}
\end{center}
\end{figure*}

More importantly, this approach helps assess the ability to train a model on a given type of search landscape (either 1-peak, 4-peak, or combined) to determine situations where classifying AI may be easier or harder. While a train-test-validation split would yield a fair model-to-model comparison, it could also lead to a biased estimation of model performance on the dataset. As a result, we utilized averaged random train-test splits to better assess data formulation quality.

\section{Experiments and Analyses}

\begin{table*}[h]
\begin{tabularx}{\textwidth}{@{}lYYYYYYYYYYYY@{}}
\toprule
&\multicolumn{3}{c}{\bfseries sharpIM}
&\multicolumn{3}{c}{\bfseries smoothIM}
&\multicolumn{3}{c}{\bfseries bmcIM}
&\multicolumn{3}{c}{\bfseries cmcIM} \\
\cmidrule(lr){2-4} \cmidrule(l){5-7} \cmidrule(l){8-10} \cmidrule(l){11-13} 
& x1 & x4 & all & x1 & x4 & all & x1 & x4 & all & x1 & x4 & all\\
\midrule
\vspace{-.25mm}
LeNet-5 & 73.3 & \textbf{76.8} & 79.6 & 73.9 & 74.7 & 77.9 & 72.3 & 74.0 & 78.1 & \textbf{75.7} & 75.7 & \textbf{79.6}\\
& (3.3) & \textbf{(3.6)} & (2.4) & (3.5) & (3.2) & (2.2) & (3.7) & (3.6) & (2.7) & \textbf{(3.4)} & (2.9) & \textbf{(2.2)} \\
\hdashline
\vspace{-.25mm}
ResNet-18 & 68.9 & \textbf{73.9} & 74.2 & 69.2 & 70.9 & 74.1 & \textbf{71.4} & 73.2 & \textbf{76.2} & 70.9 & 72.1 & 75.8\\
& (3.6) & \textbf{(3.3)} & (2.7) & (3.4) & (3.4) & (2.5) & \textbf{(3.6)} & (3.3) & \textbf{(2.5)} & (3.6) & (3.9) & (2.5) \\
\hdashline
\vspace{-.25mm}
SB-ResNet-18 & 78.7 & \textbf{82.0} & 82.3 & 77.1 & 80.2 & 81.1 & 78.7 & 79.6 & 82.0 & \textbf{78.9} & 80.4 & \textbf{82.7}\\
& (3.7) & \textbf{(3.3)} & (2.2) & (3.4) & (3.3) & (2.9) & (3.6) & (3.6) & (2.7) & \textbf{(3.4)} & (3.8) & \textbf{(2.6)}\\
\hdashline
\vspace{-.25mm}
SB-ResNet-18 & 78.7 & \textbf{82.6} & 82.8 & 78.1 & 80.7 & 81.8 & 80.2 & 80.2 & 82.8 & \textbf{80.4} & 81.5 & \textbf{83.6}\\
+ LSTM & (3.4) & \textbf{(2.9)} & (2.6) & (3.4) & (3.7) & (2.1) & (3.2) & (2.8) & (2.7) & \textbf{(3.1)} & (3.1) & \textbf{(2.6)}\\
\bottomrule
\end{tabularx}
\caption{Testing accuracy for three models averaged epoch-wise with random 80/20 splits over 100 runs on all image formulations and varied data splits. x1 was tested solely on the 1-peak environment, x4 was tested solely on the 4-peak environment, and \textit{all} was tested on the entire dataset. The bolded results are the best-performing architecture on a specific dataset. Values rounded to three significant figures to save space. }
\label{table:mr2}
\end{table*}

The highest accuracy and lowest loss were consistently observed when testing on the entire dataset (Table \ref{table:mr2}), as opposed to a subset such as solely the 1-peak or 4-peak environments, indicating the importance of accumulating more data. The combination of single block ResNet-18 and cmcIM image formulation (with supplemental LSTM) yielded the highest testing accuracy on the random split performance when averaged over 100 trials. This improvement over classical ResNet-18 and LeNet-5 indicates a balance between overfitting (ResNet-18) and underfitting (LeNet-5), achieving better testing results at the “critical regime” and optimizing the bias-variance trade-off \cite{belkin2019biasvariance,nakkiran2019dd}. The additional temporal data led to a more robust model that achieved higher test accuracy across all image formulations and datasets, giving credence to the importance of temporal data encoding.  

Loss for LeNet-5 and SB-ResNet-18 bottomed out well before training accuracy reached 100\% (20 and 10 epochs, respectively; figure \ref{figure:results}), and from there, 
continued to overfit. Results obtained on ResNet-18, however, are indicative of epoch-wise double descent, experiencing a decrease in loss and increase in testing accuracy despite over-training 
\cite{belkin2019biasvariance,nakkiran2019dd}. Despite the existence of epoch-wise double descent, ResNet-18 was still the worst-performing model due to the size of the model. 


SharpIM performed the strongest on all instances of the 4-peak landscape. 
We suspect this is due to either 1) optimal decision making on such a landscape being less dependent on exploration and exploitation decisions, so the additional channels/smoothing effects only add noise into our model inputs or 2) the 4-peaked landscapes make it easier to recognize the AI's behavior. Unfortunately, we cannot test either hypothesis without collecting new human-subject data. Both bmcIM and cmcIM obtained better performance on the 1-peak landscape due to the exploitation and exploration channels stratifying AI and non-AI movement, along with the less complex task benefitting from additional data.

Performance was strongest on the 4-peak vs 1-peak environment for most model architectures and dataset formulations. Due to the complex terrain, humans may find it difficult to select optimal dial settings. The AI, however, results in an approach that is easier to learn (e.g., following a gradient). The only instance where this property fails is in LeNet-5 and the cmcIM image formulation, where it appears that the under-parameterized model could not learn these nuances. Nevertheless, despite the smaller dataset and the difficulty detecting AI aid in abstract tasks, we achieved performance significantly better than random chance. 

For optimal model performance, we applied hyperparameter optimization for each model on the best image formulation of the combined dataset, which is included in the hyperparameter tuning section. For most models, performance between tuned and unturned parameters remained somewhat similar, likely due to the close proximity between initially selected and fine-tuned hyperparameters.

\subsection{Hyperparameter Tuning}

While we test 1-peak, 4-peak, and \textit{all} data combined, we perform hyperparameter tuning only on models for the entire dataset since testing accuracy was higher in these events. For each of the model architectures, we apply grid search on the best-performing dataset, sampling weight decay $\in \{0,5e-6,5e-5 \}$, learning rate $\in \{1e-5,1e-4,1e-3 \}$, and a linear LR scheduler $\in \{True, False \}$, and for the supplemental LSTM we additionally finetune dropout $\in (0, 0.9)$ after determining the optimal values for all other hyperparameters. We utilize 10-fold cross-validation across each hyperparameter combination instead of a traditional train-test-validate split due to dataset noisiness and our goal of showcasing realistic averaged performance. For ResNet-18 and SB-ResNet-18, we noticed the best performance at a LR of 1e-3 with a LR scheduler, with weight decay of 0 and 5e-6, respectively. For LeNet-5, we observed the best performance at a LR of 1e-4, no LR scheduler, and weight decay of 5e-6. We then retest each of these over 100 random runs, each with 80/20 dataset splits. We train each model independently as to avoid any potential data leakage.

Tuned results are shown in Table \ref{table:mr1}. For most models, we noticed better performance from high initial learning rates due to our dataset being hard-to-generalize, resulting in the model learning easy-to-fit patterns \cite{li2019scheduler}. Annealing the learning rate with a scheduler amplified this effect \cite{li2019scheduler}, along with minimizing variance between trials. Of all image formulations, cmcIM encodes the most information through its 5-channel setup, likely explaining why it was the best-performing dataset for lower parameter models, while bmcIM was the best for ResNet-18 due to potential overfitting on the cmcIM dataset. 

Compared to the unturned results, we noticed a stark improvement for ResNet-18 since the original LR of 1e-4 was too low, leading to overfitting. Most models experienced minor improvements in accuracy, however, due to the close proximity between the initial hyperparameters and the optimal hyperparameters. 

\begin{table}[h]
\centering
\begin{tabular}{lccc}
\toprule
Model & Dataset & Accuracy \\ \midrule
LeNet-5 & cmcIM & $79.93 \pm 2.52$   \\
ResNet-18 & bmcIM & $80.67 \pm 2.40$  \\
SB-ResNet-18 & cmcIM & $85.79 \pm 1.90$  \\
\hdashline
SB-ResNet-18 & \multirow{2}{*}{cmcIM} & \multirow{2}{*}{$\textbf{86.64} \pm \textbf{1.55}$}  \\ 
+ LSTM &  &   \\ 
\bottomrule
\end{tabular}
\caption{Test accuracy on best-observed dataset with 100 random 80/20 splits averaged epoch-wise, each trained to convergence.  }
\label{table:mr1}
\end{table}

\subsection{Discussion}
The ability to detect AI assistance is increasingly important as AI helpers grow in their ability to replicate human effort. Recent advances have drawn considerable attention to certain domains, such as natural language processing \cite{cotton2023chatting,rodriguez2022cross,mitchell2023detectgpt}. The methods used to detect help from an AI text generator rely on knowledge related to the function of the assistant. DetectGPT, for example, relies on insight into the probability distributions of human versus GPT text \cite{mitchell2023detectgpt}. Our method for detecting AI assistance on an abstract, complex task is unique because it does not require knowledge of how the AI agent functions.

A crucial element of our method is knowing how to present the data to models. Rather than relying on insight into how the AI assistant functioned, we relied on insight into how humans approach complex search tasks, which we note naturally extends to various other complex tasks. We experimented with four different formulations that varied in the degree to which they relied on insights related to such behavior. The single-block ResNet-18 with supplemental LSTM achieved the best performance on the entire dataset (86.64\% testing accuracy when tuned) when we included five additional layers related to exploration behavior (cmcIM). This formulation not only has indicators of whether locations were visited but also how distant those locations were from previously visited locations (in spatial and temporal domains) \textit{and} how the participant arrived at that location (was it a single-dial move or did it require both). Research on how humans conduct searches during complex tasks suggests that these details can inform how the person was thinking about the problem \cite{gurney2023aiTeammate,billinger2014search}. Theoretically, including these features in the data allowed the model to identify differences in how a given search was conducted. 

The AI assistant's behavior is embedded in the search data. As outlined in \cite{gurney2023role}, the AI assistant used a modified version of the classic simulated annealing algorithm to make its tuning decisions. Surprisingly, this feature of the AI resulted in data that looked markedly similar to human-only effort---so much so that even trained human observers cannot tell the difference. Nevertheless, the single-block ResNet-18 model with supplemental LSTM, without any inbuilt knowledge of how the AI assistant completed the task, was able to identify whether AI assistance was used more than 86\% of the time. This outcome demonstrates the possibility of detecting AI assistance with little insight into how humans typically approach abstract tasks and minimal knowledge of choice data. Knowledge of how the AI functioned was not necessary.

\section{Conclusions}
Identifying if or when humans received assistance from an AI is an increasingly important capability. Natural language agents, for example, are threatening to upend education by helping students author essays, solve word problems, and write programs \cite{firat2023chat,katz2023gpt,nori2023capabilities}. Similarly, instances of needing to determine the role of a driving assistant in a wreck are just around the corner \cite{collingwood2017privacy,westbrook2017google}. The potential need for detection is arguably only limited by humans' ability to imagine different uses for AI assistants. Problematically, each of these helpers will likely be different, and building specific detection systems will quickly become an intractable problem. Fortunately, as we have demonstrated, detection is still possible by simply relying on insights into how humans approach general classes of problems and enriching datasets with related features. 


%
%
%
\bibliographystyle{splncs04}
\bibliography{mybibliography}

\begin{thebibliography}{10}
\providecommand{\url}[1]{\texttt{#1}}
\providecommand{\urlprefix}{URL }
\providecommand{\doi}[1]{https://doi.org/#1}

\bibitem{belkin2019biasvariance}
Belkin, M., Hsu, D., Ma, S., Mandal, S.: Reconciling modern machine-learning practice and the classical bias–variance trade-off. Proceedings of the National Academy of Sciences  \textbf{116}(32),  15849--15854 (2019). \doi{10.1073/pnas.1903070116}, \url{https://www.pnas.org/doi/abs/10.1073/pnas.1903070116}

\bibitem{billinger2014search}
Billinger, S., Stieglitz, N., Schumacher, T.R.: Search on rugged landscapes: An experimental study. Organization Science  \textbf{25}(1),  93--108 (2014)

\bibitem{cai21graphnorm}
Cai, T., Luo, S., Xu, K., He, D., Liu, T.Y., Wang, L.: Graphnorm: A principled approach to accelerating graph neural network training (2021)

\bibitem{campbell2002deep}
Campbell, M., Hoane~Jr, A.J., Hsu, F.h.: Deep blue. Artificial intelligence  \textbf{134}(1-2),  57--83 (2002)

\bibitem{collingwood2017privacy}
Collingwood, L.: Privacy implications and liability issues of autonomous vehicles. Information \& Communications Technology Law  \textbf{26}(1),  32--45 (2017)

\bibitem{cotton2023chatting}
Cotton, D.R., Cotton, P.A., Shipway, J.R.: Chatting and cheating: Ensuring academic integrity in the era of chatgpt. Innovations in Education and Teaching International pp. 1--12 (2023)

\bibitem{van2023chatgpt}
van Dis, E.A., Bollen, J., Zuidema, W., van Rooij, R., Bockting, C.L.: Chatgpt: five priorities for research. Nature  \textbf{614}(7947),  224--226 (2023)

\bibitem{ezencan2020comparison}
Ezen-Can, A.: A comparison of lstm and bert for small corpus (2020)

\bibitem{firat2023chat}
Firat, M.: How chat gpt can transform autodidactic experiences and open education. Department of Distance Education, Open Education Faculty, Anadolu Unive  (2023)

\bibitem{floridi2020gpt}
Floridi, L., Chiriatti, M.: Gpt-3: Its nature, scope, limits, and consequences. Minds and Machines  \textbf{30},  681--694 (2020)

\bibitem{garbin2020bnvsdrop}
Garbin, C., Zhu, X., Marques, O.: Dropout vs. batch normalization: an empirical study of their impact to deep learning. Multimedia Tools and Applications  \textbf{79},  1--39 (05 2020). \doi{10.1007/s11042-019-08453-9}

\bibitem{philipp2018gradients}
George~Philipp, Dawn~Song, J.G.C.: Gradients explode - deep networks are shallow - resnet explained (2018), \url{https://openreview.net/forum?id=HkpYwMZRb}

\bibitem{Goodfellow-et-al-2016}
Goodfellow, I., Bengio, Y., Courville, A.: Deep Learning. MIT Press (2016), \url{http://www.deeplearningbook.org}

\bibitem{gurney2022experimental}
Gurney, N., King, T., Miller, J.H.: An experimental method for studying complex choices. In: HCI International 2022--Late Breaking Posters: 24th International Conference on Human-Computer Interaction, HCII 2022, Virtual Event, June 26--July 1, 2022, Proceedings, Part I. pp. 39--45. Springer (2022)

\bibitem{gurney2023role}
Gurney, N., Miller, J., Pynadath, D.: The role of heuristics and biases in complex choices. PREPRINT (Version 1) available at Research Square  (2023). \doi{10.21203/rs.3.rs-2472194/v1}

\bibitem{gurney2023aiTeammate}
Gurney, N., Miller, J.H., Pynadath, D.V.: The role of heuristics and biases during complex choices with an ai teammate. Proceedings of the AAAI Conference on Artificial Intelligence  \textbf{37}(5),  5993--6001 (Jun 2023). \doi{10.1609/aaai.v37i5.25741}, \url{https://ojs.aaai.org/index.php/AAAI/article/view/25741}

\bibitem{hannun2019cardiologist}
Hannun, A.Y., Rajpurkar, P., Haghpanahi, M., Tison, G.H., Bourn, C., Turakhia, M.P., Ng, A.Y.: Cardiologist-level arrhythmia detection and classification in ambulatory electrocardiograms using a deep neural network. Nature medicine  \textbf{25}(1),  65--69 (2019)

\bibitem{he2015resnet}
He, K., Zhang, X., Ren, S., Sun, J.: Deep residual learning for image recognition (2015). \doi{10.48550/ARXIV.1512.03385}, \url{https://arxiv.org/abs/1512.03385}

\bibitem{hochreiter1997lstm}
Hochreiter, S., Schmidhuber, J.: Long short-term memory. Neural Computation  \textbf{9}(8),  1735--1780 (1997). \doi{10.1162/neco.1997.9.8.1735}

\bibitem{katz2023gpt}
Katz, D.M., Bommarito, M.J., Gao, S., Arredondo, P.: Gpt-4 passes the bar exam. Available at SSRN 4389233  (2023)

\bibitem{kingma2014arxiv}
Kingma, D.P., Ba, J.: Adam: A method for stochastic optimization (2014). \doi{10.48550/ARXIV.1412.6980}, \url{https://arxiv.org/abs/1412.6980}

\bibitem{kyriakides2020nas}
Kyriakides, G., Margaritis, K.G.: An introduction to neural architecture search for convolutional networks. CoRR  \textbf{abs/2005.11074} (2020), \url{https://arxiv.org/abs/2005.11074}

\bibitem{lecun1998lenet}
Lecun, Y., Bottou, L., Bengio, Y., Haffner, P.: Gradient-based learning applied to document recognition. Proceedings of the IEEE  \textbf{86}(11),  2278--2324 (1998). \doi{10.1109/5.726791}

\bibitem{leonardos2022exploration}
Leonardos, S., Piliouras, G.: Exploration-exploitation in multi-agent learning: Catastrophe theory meets game theory. Artificial Intelligence  \textbf{304},  103653 (2022)

\bibitem{li2018visualizing}
Li, H., Xu, Z., Taylor, G., Studer, C., Goldstein, T.: Visualizing the loss landscape of neural nets (2018)

\bibitem{li2019scheduler}
Li, Y., Wei, C., Ma, T.: Towards explaining the regularization effect of initial large learning rate in training neural networks. Advances in Neural Information Processing Systems  \textbf{32} (2019)

\bibitem{mitchell2023detectgpt}
Mitchell, E., Lee, Y., Khazatsky, A., Manning, C.D., Finn, C.: Detectgpt: Zero-shot machine-generated text detection using probability curvature. arXiv preprint arXiv:2301.11305  (2023)

\bibitem{murthy2024rex}
Murthy, R., Heinecke, S., Niebles, J.C., Liu, Z., Xue, L., Yao, W., Feng, Y., Chen, Z., Gokul, A., Arpit, D., Xu, R., Mui, P., Wang, H., Xiong, C., Savarese, S.: Rex: Rapid exploration and exploitation for ai agents (2024)

\bibitem{nakkiran2019dd}
Nakkiran, P., Kaplun, G., Bansal, Y., Yang, T., Barak, B., Sutskever, I.: Deep double descent: Where bigger models and more data hurt. CoRR  \textbf{abs/1912.02292} (2019), \url{http://arxiv.org/abs/1912.02292}

\bibitem{nishihata23hai}
Nishihata, C., Kobayashi, H., Yasuda, T.: Human-like “agents” or “tools”?: Exploring the implicature-of-quantity in hai. In: Proceedings of the 11th International Conference on Human-Agent Interaction. p. 387–389. HAI '23, Association for Computing Machinery, New York, NY, USA (2023). \doi{10.1145/3623809.3623934}, \url{https://doi.org/10.1145/3623809.3623934}

\bibitem{nori2023capabilities}
Nori, H., King, N., McKinney, S.M., Carignan, D., Horvitz, E.: Capabilities of gpt-4 on medical challenge problems. arXiv preprint arXiv:2303.13375  (2023)

\bibitem{olah2015lstmvis}
Olah, C.: Understanding lstm networks  (2015)

\bibitem{paszke2019pytorch}
Paszke, A., Gross, S., Massa, F., Lerer, A., Bradbury, J., Chanan, G., Killeen, T., Lin, Z., Gimelshein, N., Antiga, L., Desmaison, A., Köpf, A., Yang, E., DeVito, Z., Raison, M., Tejani, A., Chilamkurthy, S., Steiner, B., Fang, L., Bai, J., Chintala, S.: Pytorch: An imperative style, high-performance deep learning library (2019)

\bibitem{patil22hai}
Patil, G., Bagala, P., Nalepka, P., Kallen, R.W., Richardson, M.J.: Evaluating human-artificial agent decision congruence in a coordinated action task. In: Proceedings of the 10th International Conference on Human-Agent Interaction. p. 327–329. HAI '22, Association for Computing Machinery, New York, NY, USA (2022). \doi{10.1145/3527188.3563923}, \url{https://doi.org/10.1145/3527188.3563923}

\bibitem{Ponti2022aiaid}
Ponti, M., Seredko, A.: Human-machine-learning integration and task allocation in citizen science. Humanities and Social Sciences Communications  \textbf{9}(1) (Feb 2022). \doi{10.1057/s41599-022-01049-z}, \url{https://doi.org/10.1057/s41599-022-01049-z}

\bibitem{rawat2017dcnn}
Rawat, W., Wang, Z.: Deep convolutional neural networks for image classification: A comprehensive review. Neural Computation  \textbf{29}(9),  2352--2449 (2017). \doi{10.1162/neco_a_00990}

\bibitem{rodriguez2022cross}
Rodriguez, J., Hay, T., Gros, D., Shamsi, Z., Srinivasan, R.: Cross-domain detection of gpt-2-generated technical text. In: Proceedings of the 2022 Conference of the North American Chapter of the Association for Computational Linguistics: Human Language Technologies. pp. 1213--1233 (2022)

\bibitem{schrittwieser2020mastering}
Schrittwieser, J., Antonoglou, I., Hubert, T., Simonyan, K., Sifre, L., Schmitt, S., Guez, A., Lockhart, E., Hassabis, D., Graepel, T., et~al.: Mastering atari, go, chess and shogi by planning with a learned model. Nature  \textbf{588}(7839),  604--609 (2020)

\bibitem{shamir2018resnets}
Shamir, O.: Are resnets provably better than linear predictors? (2018)

\bibitem{silver2016mastering}
Silver, D., Huang, A., Maddison, C.J., Guez, A., Sifre, L., Van Den~Driessche, G., Schrittwieser, J., Antonoglou, I., Panneershelvam, V., Lanctot, M., et~al.: Mastering the game of go with deep neural networks and tree search. nature  \textbf{529}(7587),  484--489 (2016)

\bibitem{silver2017mastering}
Silver, D., Schrittwieser, J., Simonyan, K., Antonoglou, I., Huang, A., Guez, A., Hubert, T., Baker, L., Lai, M., Bolton, A., et~al.: Mastering the game of go without human knowledge. nature  \textbf{550}(7676),  354--359 (2017)

\bibitem{thorp2023chatgpt}
Thorp, H.H.: Chatgpt is fun, but not an author (2023)

\bibitem{turing1950computing}
Turing, A.: Computing machinery and intelligence-am turing. Mind  \textbf{59}(236), ~433 (1950)

\bibitem{uchendu2023understanding}
Uchendu, A., Lee, J., Shen, H., Le, T., Huang, T.H., Lee, D.: Understanding individual and team-based human factors in detecting deepfake texts. arXiv preprint arXiv:2304.01002  (2023)

\bibitem{vaswani2023attention}
Vaswani, A., Shazeer, N., Parmar, N., Uszkoreit, J., Jones, L., Gomez, A.N., Kaiser, L., Polosukhin, I.: Attention is all you need (2023)

\bibitem{vinyals2019grandmaster}
Vinyals, O., Babuschkin, I., Czarnecki, W.M., Mathieu, M., Dudzik, A., Chung, J., Choi, D.H., Powell, R., Ewalds, T., Georgiev, P., et~al.: Grandmaster level in starcraft ii using multi-agent reinforcement learning. Nature  \textbf{575}(7782),  350--354 (2019)

\bibitem{wang2022adversarial}
Wang, T.T., Gleave, A., Belrose, N., Tseng, T., Miller, J., Dennis, M.D., Duan, Y., Pogrebniak, V., Levine, S., Russell, S.: Adversarial policies beat professional-level go ais. arXiv preprint arXiv:2211.00241  (2022)

\bibitem{weizenbaum1966eliza}
Weizenbaum, J.: Eliza—a computer program for the study of natural language communication between man and machine. Communications of the ACM  \textbf{9}(1),  36--45 (1966)

\bibitem{westbrook2017google}
Westbrook, C.W.: The google made me do it: the complexity of criminal liability in the age of autonomous vehicles. Mich. St. L. Rev. p.~97 (2017)

\bibitem{wilson2021balancing}
Wilson, R.C., Bonawitz, E., Costa, V.D., Ebitz, R.B.: Balancing exploration and exploitation with information and randomization. Current opinion in behavioral sciences  \textbf{38},  49--56 (2021)

\bibitem{xu2018how}
Xu, K., Hu, W., Leskovec, J., Jegelka, S.: How powerful are graph neural networks? In: International Conference on Learning Representations (2019), \url{https://openreview.net/forum?id=ryGs6iA5Km}

\bibitem{Yao2019cnnrnn}
Yao, H., Zhang, X., Zhou, X., Liu, S.: Parallel structure deep neural network using cnn and rnn with an attention mechanism for breast cancer histology image classification. Cancers  \textbf{11}(12), ~1901 (Nov 2019). \doi{10.3390/cancers11121901}, \url{http://dx.doi.org/10.3390/cancers11121901}

\bibitem{yun2019resnetsbetter}
Yun, C., Sra, S., Jadbabaie, A.: Are deep resnets provably better than linear predictors? CoRR  \textbf{abs/1907.03922} (2019), \url{http://arxiv.org/abs/1907.03922}

\bibitem{zhang2020adamvssgd}
Zhang, J., Karimireddy, S.P., Veit, A., Kim, S., Reddi, S.J., Kumar, S., Sra, S.: Why {\{}adam{\}} beats {\{}sgd{\}} for attention models (2020), \url{https://openreview.net/forum?id=SJx37TEtDH}

\end{thebibliography}
%





\end{document}